\documentclass[10pt,twocolumn,letterpaper]{article}

\usepackage[pagenumbers]{cvpr} 

\usepackage{graphicx}
\usepackage{caption}
\usepackage{cuted}
\usepackage{amsmath}
\usepackage{mdframed}
\usepackage{multirow}  %
\usepackage{makecell}
\usepackage[dvipsnames]{xcolor}
\usepackage{makecell}
\usepackage{array}

\definecolor{cvprblue}{rgb}{0.21,0.49,0.74}
\usepackage[pagebackref,breaklinks,colorlinks,citecolor=cvprblue]{hyperref}
\usepackage{float}

\def\paperID{14677} 
\def\confName{CVPR}
\def\confYear{2025}

\title{The Language of Motion: \\
Unifying Verbal and Non-verbal Language of 3D Human Motion}

\author{
Changan Chen$^*$ \hspace{1mm} Juze Zhang$^*$ \hspace{1mm} Shrinidhi K. Lakshmikanth$^*$ \hspace{1mm} Yusu Fang \hspace{1mm} Ruizhi Shao \hspace{1mm}   \\
Gordon Wetzstein \hspace{1mm}
Li Fei-Fei \hspace{1mm} Ehsan Adeli \hspace{1mm}
\\
Stanford University
}

\begin{document}
\maketitle

\def\thefootnote{*}\footnotetext{indicates equal contribution}

\begin{strip}\centering
    \vspace{-50px}
    \captionsetup{type=figure}
    \includegraphics[width=1\textwidth]{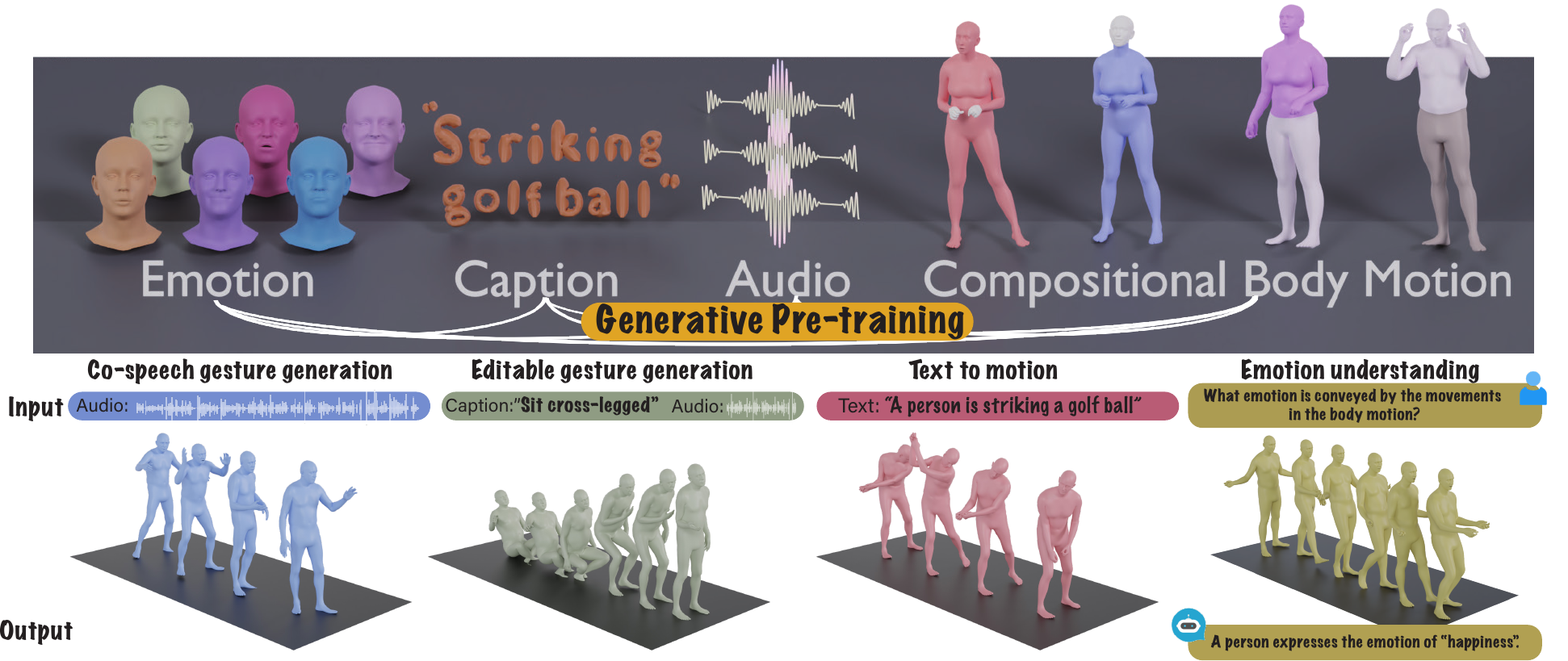}
    \vspace{-15px}
    \captionof{figure}{We introduce a language-model-based motion understanding and generation framework that takes in any of the audio/motion/text modalities and outputs the desired target modality. Coupled with our generative pre-training strategy, our model demonstrates competitive performance on an array of tasks, showing promising signs toward unified verbal and non-verbal language of human motions.
    }
    \label{fig:teaser}
\end{strip}

\begin{abstract}
Human communication is inherently multimodal, involving a combination of verbal and non-verbal cues such as speech, facial expressions, and body gestures. Modeling these behaviors is essential for understanding human interaction and for creating virtual characters that can communicate naturally in applications like games, films, and virtual reality.  However, existing motion generation models are typically limited to specific input modalities—either speech, text, or motion data—and cannot fully leverage the diversity of available data. In this paper, we propose a novel framework that unifies verbal and non-verbal language using multimodal language models for human motion understanding and generation. This model is flexible in taking text, speech, and motion or any combination of them as input. Coupled with our novel pre-training strategy, our model not only achieves state-of-the-art performance on co-speech gesture generation but also requires much less data for training. Our model also unlocks an array of novel tasks such as editable gesture generation and emotion prediction from motion. We believe unifying the verbal and non-verbal language of human motion is essential for real-world applications, and language models offer a powerful approach to achieving this goal. Project page: \url{languageofmotion.github.io}.
\end{abstract}  
    
\section{Introduction}
\label{sec:intro}


Human communication is multimodal. We use spoken and body language, including hand gestures, facial expressions, body postures and emotional expressions to interact with each other effectively. For example, people use linguistic cues along with body language, including hand gestures, facial expressions, overall body posture, and even emotional expressions to interact with the environment effectively. Modeling these multimodal behaviors is essential for understanding and generating human motion, enabling a wide range of applications for virtual characters in games, movies, and virtual reality—areas that have recently received substantial attention.

Existing work has been focused on modeling human motion from different modalities, such as speech~\cite{liu24emage,yi23generating,ng24audio}, text~\cite{zhang23t2mgpt,jiang23motiongpt,wu24motionllm}, egocentric vision~\cite{li2024egogen,li2023ego}, or the surrounding environment~\cite{jiang2024scaling,wang2024move,zhang2024hoim3, zhang2023neuraldome, bhatnagar2022behave}. These models only take specific modalities as input, whose performance is thus limited to the data available for their downstream task. For example, co-speech gesture generation work typically trains speaker-dependent models~\cite{habibie21learning,yi23generating,liu24emage, ao2023gesturediffuclip}, which requires high-quality speech--motion capture of a person.
While gesture style varies from person to person, many gestures are shared across people as well as non-speech-driven motion, such as walking or waving hands. Existing work has yet to leverage motion priors from all forms of motion data.




One of the promising ways to unify different tasks is multimodal language models, where a single language model can take different modalities as input and output target modalities. These models have shown promising results in a wide range of multimodal tasks, such as visual question answering~\cite{liu2023llava,li2023blip2,alayrac2022flamingo}, audio understanding and generation~\cite{audiolm,gong2023ltu,zhang2023speechgpt}, and text-to-motion generation~\cite{zhang24motiongpt,chen24motionllm,zhang23t2mgpt, wang2024motiongpt,chen2023executing}. While language models have been widely applied, it has not been explored in the speech-text--motion generation setting.


We argue language models play a crucial role in unifying the verbal and non-verbal language of human motion for three reasons: 1) language models naturally connect different modalities, 2) speech is highly semantic, and tasks like modeling laughter in response to a joke require strong semantic reasoning capabilities and 3) language models are equipped with strong semantic understanding from extensive pre-training.

Towards this goal, we propose a novel multimodal language model for expressional motion generation and understanding (see Fig.~\ref{fig:teaser}).
To leverage language models to model motion, we first tokenize motion separately for different body parts (face, hand, upper-body, lower-body). Such compositionality has shown to be more beneficial to model the expressive human expressions~\cite{ng24audio,liu24emage}. Along with off-the-shelf tokenizers for text and speech~\cite{kudo2018sentencepiece}, we can represent any given modality inputs as a sequence of tokens, which are consumed by language models. To train the language models, we design a two-stage training pipeline. The model is first pre-trained to align various modalities with compositional body motion alignment and audio--text alignment. After pre-training, we compile downstream tasks into instructions and train the model on these instructions to allow the model to follow various task instructions.


We first validate our model on the BEATv2 co-speech gesture generation benchmark~\cite{liu24emage} and show that our model strongly outperforms state-of-the-art models. We then conduct a thorough evaluation to demonstrate the effectiveness of our pre-training tasks. We also show that our pre-training strategy is more powerful when under severe data scarcity. While never seeing speech--motion data during pre-training, our model reaches competitive performance with a relatively small amount of data for a novel speaker, showing remarkable generalization. By performing post-training on both speech--motion and text--motion tasks, we show that our model not only follows audio and text prompts but also unlocks novel tasks such as predicting emotion from motion data. Please watch the Supp. video for the qualitative examples. To the best of our knowledge, this is the first work to build multimodal language models to unify the verbal and non-verbal language of 3D human motions.


\section{Related Work}
\label{sec:related_work}

\begin{figure*}[t]
    \centering
    \includegraphics[width=0.9\linewidth]{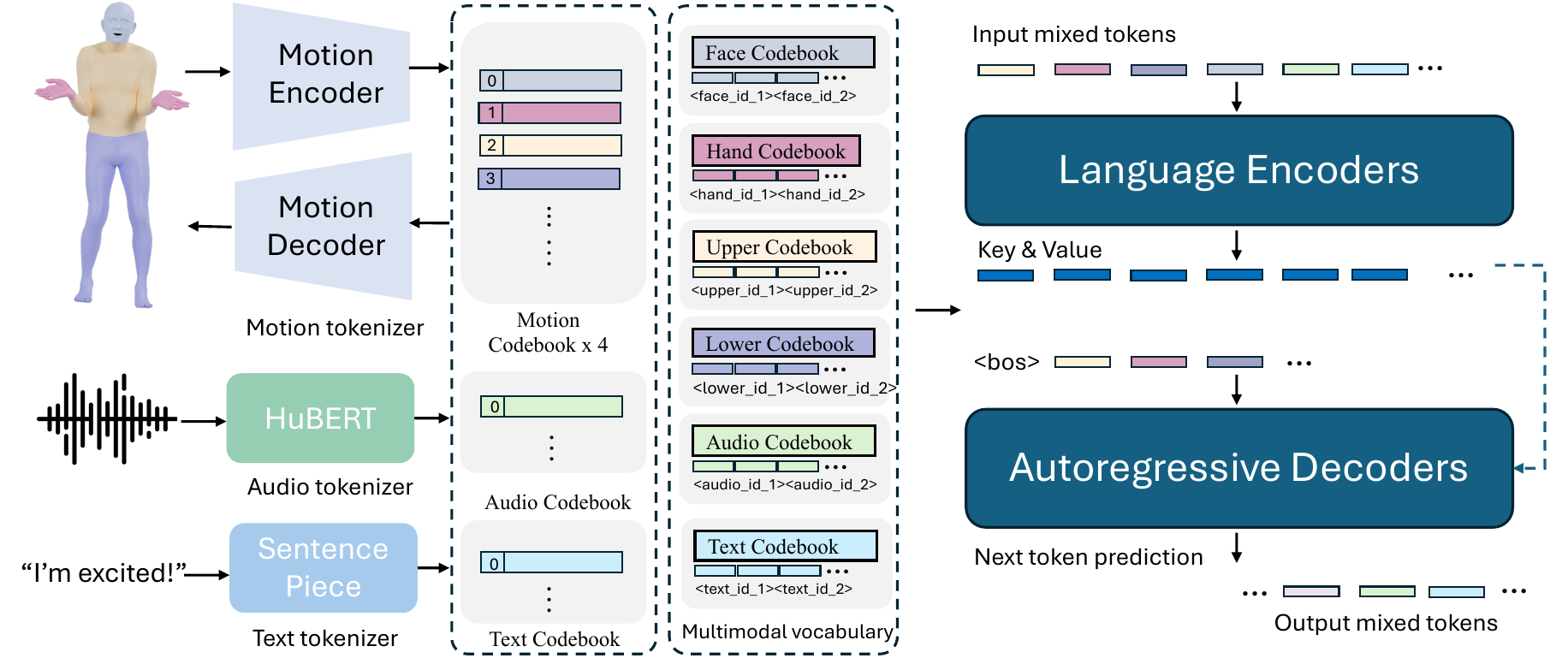}
    \vspace{-0.05in}
    \caption{Method overview. We employ modality-specific tokenizers to process various input modalities. Specifically, we train a compositional body motion VQ-VAE to tokenize face, hands, upper body, and lower body motions into discrete tokens, combining these modality-specific vocabularies(audio and text) into a unified multimodal vocabulary. During training, mixed tokens from different modalities are used as input, and the output is generated through an encoder-decoder language model. The mixed tokens are fed into the transformer encoder, while the decoder predicts the probability distribution of the next token in an autoregressive manner at each step.
    }
    \label{fig:approach}
    \vspace{-0.15in}
\end{figure*}

\subsection{Speech-Driven Motion Generation}
Human communication is multimodal and we use our speech, facial expressions, and body gestures to communicate with each other. Given this complimentary nature, recent work~\cite{ahuja2020style,li2021audio2gestures} explores cross-modal generation of human motion from speech in different forms.
These models are trained and evaluated on specific upper body joints, full body joints, and even facial expressions. Recent work in co-speech gesture generation often utilizes generative models to create gestures from audio conditions~\cite{Chhatre_2024_CVPR,yi2023generating,chen2024Synerg,liu24emage}. Other work also explores the possibility of generating the listener's motion~\cite{ng22learning,ng23can,tran24dyadic}. 
Another area of research focuses on generating speech-driven co-speech facial expressions, with notable works including ViCO~\cite{zhou2022responsive} and CodeTalker~\cite{xing2023codetalker}. 
However, these works are limited in that they only take speech as input and do not utilize other forms of motion data, making it challenging to follow both speech and textual cues. To tackle this, we propose to unify input/output modalities with a language model framework.

\subsection{Text to Motion Generation}
Humans communicate with spoken language and non-verbal means such as emotions and interactions with surrounding environment~\cite{jiang2024scaling,wang2024move}, among other cues.
Recent work has explored generating human motion from text descriptions~\cite{ guo2024momask, liu2023plan, zhang2023generating, jiang2025motionchain, wang2023intercontrol, guo2022generating, petrovich2022temos, shafir2023human, xie2023omnicontrol, karunratanakul2023guided, zhang2025freemotion, zhang2023generating, athanasiou2024motionfix, chi2024m2d2m, guo2022tm2t}. Some work attempts to generate human motion using diffusion models~\cite{tevet22human,chen23executing,zhang2022motiondiffuse,zhang2022motiondiffuse, tevet2023human, zhou2025emdm, yuan2023physdiff, zhang2023remodiffuse} while other work exploring language models for generating human motion~\cite{zhang23t2mgpt,jiang23motiongpt,wu24motionllm, zhang24motiongpt, chen24motionllm, wang2024motiongpt, liang2024omg,zhang2024large, wu24motionllm}.
While these works have shown promising results in generating human motion from text instructions, they fall short in capturing the underlying meaning of the motion language itself. This limitation makes it challenging to develop a model capable of generating human motion from both verbal and non-verbal language. In this work, we propose a novel framework to capture patterns in body language and subtle expressive gestures inherently present in human communication.

\subsection{Multimodal Language Models}
Recent years have witnessed the rise of language models~\cite{zhao2024survey,bert,gpt3,t5, wu2024vila, po2024state}, primarily leveraging transformer architectures~\cite{vaswani17attention} that process text tokens as input and generate text tokens.
Building upon these advancements, substantial efforts have expanded into multimodal language models capable of handling various types of input and output, with notable examples including BLIP-2~\cite{li2023blip}, LLaVA~\cite{liu2023llava}, and VideoChat~\cite{li2023videochat}.
Furthermore, the scope of multimodal language models (MM-LLMs) has broadened to include modality-specific outputs, as demonstrated by models like GILL~\cite{koh2024generating} and SpeechGPT~\cite{zhang2023speechgpt}. Efforts such as LLaVA~\cite{liu2023llava} and AudioGPT~\cite{huang2024audiogpt} are advancing towards seamless any-to-any modality conversion, with the goal of emulating human-like cognitive abilities in multimodal contexts.
Inspired by this line of work, we propose a new framework aimed at unifying verbal and non-verbal language within language models. 
Our framework takes text, speech, and motion data as input and generates human motion or text as output, further exploring the potential synergy between different tasks and modalities to enhance the performance of human motion generation.


\section{Multimodal Language Model for Motion Generation and Understanding}
\label{sec:approach}
In this section, we present a multimodal language model for motion generation and understanding, which is illustrated in Fig.~\ref{fig:approach}. We first describe the tokenization of different modalities (Sec.~\ref{Tokenization}), then we introduce our generative pre-training for modality alignment (Sec.~\ref{prtraining}), and finally, we detail post-training for instructions following(Sec.~\ref{Posttraining}).

\subsection{Preliminaries}

We use the neutral SMPL-X \cite{pavlakos2019smplx} body model including FLAME \cite{li2017learning} face model. This model is parameterized by per-person body shape $\mathbf{\beta} \in \mathbb{R}^{T \times 300}$, 55 joint pose $ \mathbf{g} \in \mathbb{R}^{T \times 55 \times 3} $, facial expression $\mathbf{\psi} \in \mathbb{R}^{T \times 100} $, and global body translation $\mathbf{\gamma}\in\mathbb{R}^{T \times 3}$, where $T$  is the frame number.


\subsection{Tokenization}
\label{Tokenization}
In order for our modelt  take various modalities as input (audio, tex,t and motion),  we first tokenize different modalities with modality-specific tokenizers, and then combine them into a multimodal vocabulary.

\noindent\textbf{Compositional body motion tokenization.}
Following the motion representation approach in EMAGE~\cite{liu24emage}, we divide the body into four parts with 6D rotation representation: 9 joints forming the lower-body $\mathbf{g}_l \in \mathbb{R}^{T\times54}$, 13 joints forming the upper-body $\mathbf{g}_u \in \mathbb{R}^{T\times78}$, 30 joints forming the hands $\mathbf{g}_h \in \mathbb{R}^{T\times180}$ and 1-joint along with 100 expression parameters representing the face $\mathbf{g}_f \in \mathbb{R}^{T\times106}$. Collectively, the motion space is represented as $G = \{\mathbf{g}_f, \mathbf{g}_h, \mathbf{g}_u, \mathbf{g}_l \}$.  
Notably, we avoid using the commonly adopted HumanML3D representation \cite{guo2022generating} (H3D-Format) in text-to-motion tasks, as it predominantly focuses on skeletal movement, emphasizing swinging motions while overlooking twisting rotations of body parts—an essential aspect for effectively conveying body language.
With this compositional representation, we train four separate VQ-VAEs to tokenize the body pose for each part. Each VQ-VAE encoder $\mathcal{E}$ applies a four-layer  temporal convolutional network (TCN) to extract continuous latent motion features $ {\mathbf{z}}^{1:T} = \mathcal{E}(\mathbf{g}^{1:T}) $. This encoded representation $ \mathbf{z}^{1:T} $ is quantized using:
\begin{align}
\begin{split}    
    \mathbf{q}^t = \mathcal{Q} ( \mathbf{z}^t ) := \arg\min_{\mathbf{q}^k \in Q} \| \mathbf{z}^t - \mathbf{q}^k \|^2 
\end{split}
\end{align}
where $\mathbf{q}^t$ is the discrete code in the codebook representing the encoded $\mathbf{z}^t$. Collectively, the quantized motion latent space is $ Q=\{ \mathbf{q}_f, \mathbf{q}_h, \mathbf{q}_u, \mathbf{q}_l \} $. Each VQ-VAE decoder $\mathcal{D}$ decodes the quantized motion $\mathbf{q}^t$ back into the motion space $ \mathbf{\hat{g}}^{1:T} = \mathcal{D}(\mathbf{q}^{1:T}) $ and applies the following reconstruction losses:
\begin{align}
\begin{split}    
    \mathcal{L}_{total} =& 
    \mathcal{L}_{\text{rec}} (\mathbf{g},\mathbf{\hat{g}}) + 
    \mathcal{L}_{\text{vel}} (\mathbf{g'},\mathbf{\hat{g}'}) + \\
    &
    \mathcal{L}_{\text{acc}} (\mathbf{g''},\mathbf{\hat{g}''}) +
    \mathcal{L}_{\text{mrec}} (\mathbf{g},\mathbf{\hat{g}}) + \\
    &
    \mathcal{L}_{\text{mvel}} (\mathbf{g'},\mathbf{\hat{g}'}) + 
    \mathcal{L}_{\text{macc}} (\mathbf{g''},\mathbf{\hat{g}''}) + \\
    & \mathcal{L}_{\text{comm}} (\mathbf{g},\mathbf{{q}}),
\end{split}
\end{align}
where
$\mathbf{\hat{g}'}$ represent the reconstructed motion, $\mathbf{\hat{g}'}$ and $\mathbf{g'}$ represent the velocity of $\mathbf{\hat{g}}$ and $\mathbf{g}$, while $\mathbf{\hat{g}''}$ and $\mathbf{g''}$ represent their acceleration. 
For lower-body, upper-body and hands VQ-VAEs, pose reconstruction loss $\mathcal{L}_{rec}$ is a Geodesic loss. For face VQ-VAE, $\mathcal{L}_{rec}$ is $\ell_2$ loss. Pose velocity/acceleration losses $\mathcal{L}_{vel}$ and $\mathcal{L}_{acc}$ are $\ell_1$ losses. Mesh reconstruction loss $\mathcal{L}_{mrec}$ is $\ell_2$ loss. Mesh velocity/acceleration losses $\mathcal{L}_{mvel}$ and $\mathcal{L}_{macc}$ are $\ell_1$ losses. Codebook commitment loss $\mathcal{L}_{comm}$ is $\ell_2$ loss. Vertices of the SMPLX-2020 mesh computed from the pose $\mathbf{g}$ and $\mathbf{\hat{g}}$ are used to compute mesh losses.

\noindent\textbf{Speech tokenization.}
Similar to the motion modality, speech data is also continuous by nature. To facilitate speech training within a language model, we used HuBERT \cite{hsu2021hubert} to represent audio streams as discrete tokens. In this work, audio input is sampled at 16 kHz, resulting in $ \mathbf{a} \in  \mathbb{R} ^{T \times s}$, where $s$ represents the audio frame rate after quantization. HuBERT further downsamples audio by a factor of 320, resulting in $s=50$. This frame rate, compared to the typical motion frame rate of 30 fps, provides an acceptable input token length for language models. The resulting audio token space is noted as $A = \{\mathbf{a}\}$.

\noindent\textbf{Text tokenization.}
Following previous work \cite{jiang23motiongpt,t5}, we use SentencePiece \cite{kudo2018sentencepiece} to tokenize text inputs and outputs into WordPiece tokens \cite{sennrich2015neural,kudo2018subword} for the language model, with a vocabulary of 32,000 wordpieces inherited from the T5 \cite{t5} language model, which can be represented as $W= \{\mathbf{w}\}$. This vocabulary enables the model to process a fixed set of predetermined languages. Additionally, we extend the vocabulary with several multimodal tokens to support multi-modal inputs.


\begin{figure}[t]
    \centering
    \includegraphics[width=\linewidth]{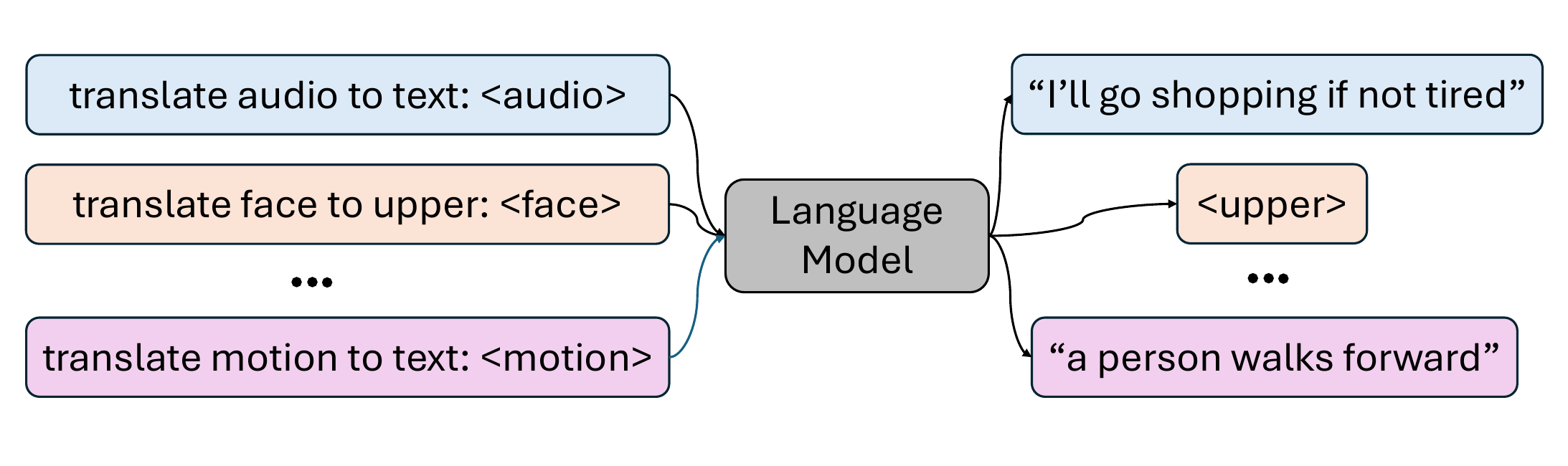}
    \vspace{-0.3in}
    \caption{Illustration of pre-training. We pre-train our language model by translating one modality to another using paired data.}\label{fig:training}
    \vspace{-0.1in}
\end{figure}

\noindent\textbf{Multimodal vocabulary.}
Altogether, we have a combined token space defined as $M:= Q \cup A \cup W \cup  C =\{ \mathbf{q}_f, \mathbf{q}_u, \mathbf{q}_h, \mathbf{q}_l, \mathbf{a}, \mathbf{w} \}$. Each modality-specific tokenizer outputs its modality-specific vocabulary. To build a unified language model that can process these different modalities, we need to combine these vocabularies into a joint vocabulary. Since the language model is pre-trained with the text modality, we choose to extend the original text vocabulary $V_t =\{ v_t^i\}_{i=1}^{K_t}$ with vocabularies from other modalities, including audio $V_a =\{ v_a^i\}_{i=1}^{K_a}$, face $V_f =\{ v_f^i\}_{i=1}^{K_f}$, hands $V_h =\{ v_h^i\}_{i=1}^{K_h}$, upper body  $V_u =\{ v_u^i\}_{i=1}^{K_u}$, and lower body $V_l =\{ v_l^i\}_{i=1}^{K_l}$, following previous work \cite{jiang23motiongpt}.
In particular, the motion vocabulary is defined as a combination of four body-part vocabularies: $V_m = \{ v_f^i, v_h^i, v_u^i, v_l^i \}_{i=1}^{K_m}$.
Additionally, each modality-specific vocabulary includes special tokens for boundary recognition, such as $<$/soa$>$  and $<$/eoa$>$ to indicate the start and end of an audio sequence. As a result, all modalities can be represented in a unified format with one joint multimodal vocabulary $V = \{ V_t, V_a, V_f, V_h, V_u, V_l \} $.

\subsection{Pre-training for Modality Alignment} 
\label{prtraining}
Existing motion generation models rely heavily on paired data to train downstream tasks. Yet, collecting high-quality paired motion data is both costly and time consuming while there exists a large amount of unpaired data of each modality that can be explored. Inspired by this, we introduce our generative pre-training strategy, as shown in Fig.~\ref{fig:training}. More specifically, we implement two types of modality alignment during the pre-training stage: compositional motion alignment and audio--text alignment that are detailed below.

\noindent\textbf{Compositional body motion alignment.}
Our body motion is inherently compositional, i.e., different body parts move in accordance. For example, when we are happy, our faces express smiles and our gestures tend to become more positive. The correlation between different body part motions is universal, transcending cultural boundaries. This shared prior forms the basis of our approach. To explore this correspondence, we consider two types of motion alignment tasks: spatial and temporal. 

\textbf{Spatial.} To model the correlation between these different body parts, we train the model to take in a randomly selected combination of body parts (e.g., upper or upper + face) and predict another randomly selected combination of other body parts (e.g., lower or lower + hand). This helps our model learn the spatial relations between body parts. 
Below is one example template that defines task prompts, conditions, and answers. The model takes both prompts and conditions as input and is expected to output the answer.

\begin{mdframed}
\textcolor{blue}{Task Prompts}: Translate upper to lower body. \\
\textcolor{blue}{Conditions:} Upper Body Tokens $ 
V_\text{condition} = \{ v_u^i \in V_u \mid i \in \{ \text{sequence token index} \} \} $ \\
\textcolor{blue}{Answer}: Lower Body Tokens $ 
V_\text{Answer} = \{ v_l^i \in V_l \mid i \in \{ \text{sequence token index} \} \} $
\end{mdframed}

\textbf{Temporal.} Predicting how motion changes as a function of time is also an important self-supervision, which enables the model to capture the temporal evolution of motion. We model this by randomly masking off certain motion frames to help the model learn the temporal priors of motion. 
\begin{mdframed}
\textcolor{blue}{Task Prompts}: Translate mask to unmasked motion. \\
\textcolor{blue}{Conditions:} Masked Tokens $ 
V_\text{condition} = \{ v_m^i \in V_m \mid i \in \{ \text{masked sequence token index} \} \}.  $\\
\textcolor{blue}{Answer}: Unmasked Motion Tokens $ 
V_\text{Answer} = \{ v_m^i \in V \mid i \in \{ \text{unmasked sequence token index} \} \} $
\end{mdframed}


\noindent\textbf{Audio-text alignment.}
In addition to the motion modality, we also design translation tasks between audio and text modalities, leveraging the abundance of available data.
These tasks follow the format of “predicting modality Y from modality X". For example, ``predicting text from audio" should help the model's performance in ``predicting motion from audio" by mapping the audio embeddings into the well-pre-trained text embedding space.


\subsection{Post-training with Instruction Following}
\label{Posttraining}

After pre-training, the model gains an understanding of the underlying grammar and syntax within motion modality’s vocabulary and good alignment between audio and text modalities. We then fine-tune the model with paired data on downstream tasks such as co-speech gesture generation or text-to-motion generation.
To enable the model to perform desired downstream tasks while following natural human instructions, 
we construct a multi-task instruction-following template by formatting several key tasks such as audio-to-motion, text-to-motion, and emotion-to-motion into instructions.
Specifically, for each task, we compose dozens of different instruction templates, resulting in more than one thousand different tasks, each having a unique instruction prompt.
An example of our instruction template is shown below. See Supp. for more examples.
\begin{mdframed}
\textcolor{blue}{Task Prompts}: Based on $<\text{Audio} \_ \text{Placeholder}>$, generate a full-body movement sequence involving face, hands, upper body, and lower body that matches the audio’s rhythm. \\
\textcolor{blue}{Conditions:} Audio Tokens $ 
V_\text{condition} = \{ v_a^i \in V_a \mid i \in \{ \text{audio sequence token index} \} \} $ \\
\textcolor{blue}{Answer}: Unmasked Motion Tokens $ 
V_\text{Answer} = \{ v_m^i \in V \mid i \in \{ \text{ motion sequence token index} \} \} $
\end{mdframed}


\begin{table*}[th]
    \centering
    \begin{tabular}{ccccccc}
        \toprule
         & FGD $\downarrow$ & BC$\uparrow$ & Diversity$\uparrow$ & Condition Signal\\
        \midrule
        DisCo~\cite{liu2022disco}  &9.417 & 6.439 & 9.912  & audio \\
        CaMN~\cite{liu22beat} & 6.644 &6.769 &10.86  & audio, text, facial\\
        DiffStyleGesture~\cite{ijcai2023p650} &8.811 &7.241 &11.49  & audio, style\\
        Habibie et al.~\cite{habibie21learning} &9.040 & 7.716 &8.213  & audio, text \\
        TalkSHOW~\cite{yi23generating} & 6.209 & 6.947 & 13.47   & audio \\
        SynTalker~\cite{chen2024Synerg} & 6.413 & 7.971 & 12.721 & audio, text\\
        EMAGE~\cite{liu24emage} & 5.512 & 7.724 &13.06  & audio, text\\
        \midrule
        Ours w/o language pre-training & 7.470 & 6.148 & 14.162 &  audio\\
        Ours w/o multimodal pre-training &  5.408 & 7.742 & 14.418   & audio\\
        Ours        &   \textbf{5.301} & \textbf{7.780} & \textbf{15.167}  & audio\\
        \bottomrule
    \end{tabular}
    \caption{Co-speech gesture generation results on BEATv2 benchmark. We report FGD $\times 10^{-1}$, BC $\times 10^{-1}$, Diversity, which measures realism, speech-motion synchronization, and diversity respectively. Our model outperforms state-of-the-art methods on this benchmark.}
    \label{tab:audio2motion}
    \vspace{-0.2in}
\end{table*}

\subsection{Language Model Training Details} 
Our model leverages 220M pre-trained Flan-T5-Base model \cite{t5} with an encoder--decoder transformer structure to address the conditional generation task. Through our shared multimodal vocabulary $V$, every input modality is represented as ``text" tokens, allowing us to fully leverage the original T5 model for each conditional generation task. Specifically, with constructed modality latent codebook indices—such as index 8 of the upper body codebook—the upper body can be formatted as ``$<\text{upper}\_8>$". Thus, the input can be converted into a sequence of tokens $S_i = \{ s_i^k \}_{k=1}^L$, where $s_i \in V$ and $L$ represents the input length. Similarly, the model outputs a sequence of tokens $S_o = \{ s_o^k \}_{k=1}^L$, with a fixed input/output token length and $s_o \in V$.

Since our model is encoder-decoder architecture, we set a maximum input length of 512. We specify modalities with start and stop tokens. Following the original T5 implementation, the sequence of tokens is sent to the encoder, and the decoder then performs next-token predictions in an auto-regressive manner at each step.
The training objective can be formulated as follows:
\begin{equation}
\mathcal{L}_{LM} = - \sum_{k=0}^{L_t-1}\log p_{\theta}(s_t^k | s_t^{<k},s_i),
\end{equation}
where $s_t$ represents each token within the sequence, serving as the index in our unified vocabulary $V$. Through next-token prediction, our model learns the underlying distribution of each modality, enabling the accurate and meaningful generation of target ``words".
For both pre-training and post-training, we finetune the model’s entire weights instead of performing low-rank adaptation (LORA~\cite{hu2021lora}) since our goal is to maximally align each modality.
\begin{table}[th]
    \centering
    \begin{tabular}{cccc}
        \toprule
         & FGD $\downarrow$ & BC$\uparrow$ & Diversity$\uparrow$ \\
        \midrule
        W/o pre-training & 5.501 & 7.721 & 14.281 \\
        W/o A2T &  5.443 & 7.721 & 14.499 \\
        W/o spatial  &  6.336 & 7.381 & 14.173 \\
        W/o temporal & 6.800  & 7.341 & 13.810 \\
        W/o motion &  7.776 & 7.344 & 14.640  \\
        Ours        &   \textbf{5.301} & \textbf{7.780} & \textbf{15.165} \\
        \bottomrule
    \end{tabular}
    \caption{Ablations of pre-training. 
    }
    \label{tab:ablation_pretrain}
    \vspace{-0.2in}
\end{table}

\section{Experiments}
\label{sec:experiments}

In this section, we first evaluate our model on the co-speech gesture generation benchmark, then investigate the generalization enabled by generative multimodal pre-training, demonstrate our model's capability in following both audio and text prompts, and lastly our novel ability to predict emotion from motion.




\subsection{Co-Speech Gesture Generation}\label{sec:audio2motion}
To evaluate our model's audio-to-motion generation ability, we choose to benchmark on co-speech gesture generation on the BEATv2 dataset~\cite{liu24emage}, where the goal is to generate body gesture motion for a given speech of a speaker. Existing co-speech generation work typically models speaker-dependent gestures~\cite{yi23generating,ng23can,liu24emage}.
During the pre-training stage, our model utilizes large-scale unpaired data in a self-supervised setup, drawing from two primary datasets, BEATv2 and Librispeech~\cite{panayotov2015librispeech}. Together, these datasets provide approximately 1,000 hours of audio-to-text data and 60 hours of motion data.
During the post-training, to ensure a fair comparison with baselines, we adopt the same evaluation protocol as~\cite{liu24emage}, i.e., training and testing on speaker-2 and using their motion tokenizer. For pre-training, we ensure the model does not see any audio-to-motion data. Following prior work~\cite{liu24emage}, we adopt Frechet Gesture Distance (\textbf{FGD})~\cite{Yoon2020Speech} to evaluate the realism of the body gestures, Beat Correlation (\textbf{BC})~\cite{li2021learn} to assess speech-motion synchrony and \textbf{Diversity}~\cite{li2021audio2gestures}, which is calculated with the $\ell_1$ distance between multiple body gesture clips.

The results are shown in Table~\ref{tab:audio2motion}. Compared with the state-of-the-art methods on this benchmark, our model achieves better performance across all metrics, indicating that our model generates more realistic, and diverse motion that is synchronized with the speech. Existing work often supplies additional signals to the model to boost the model's performance such as the text transcribed from the speech~\cite{liu22beat,habibie21learning,liu24emage} or onset/amplitudes~\cite{liu24emage}, partially due to the lack of semantic understanding of speech. In our model, since we use a pre-trained language model, the model naturally has a strong semantic understanding, allowing our model to show competitive performance without heavy reliance on hand-crafting features. If we use randomly initialized model weights, we can see that the model's performance drops drastically, indicating that language pre-training is vital for co-speech gesture generation. If we remove our multimodal pre-training stage, the model's performance also deteriorates, showing that our model benefits from the generative pre-training.

\begin{figure*}[th]
    \centering
    \includegraphics[width=\textwidth]{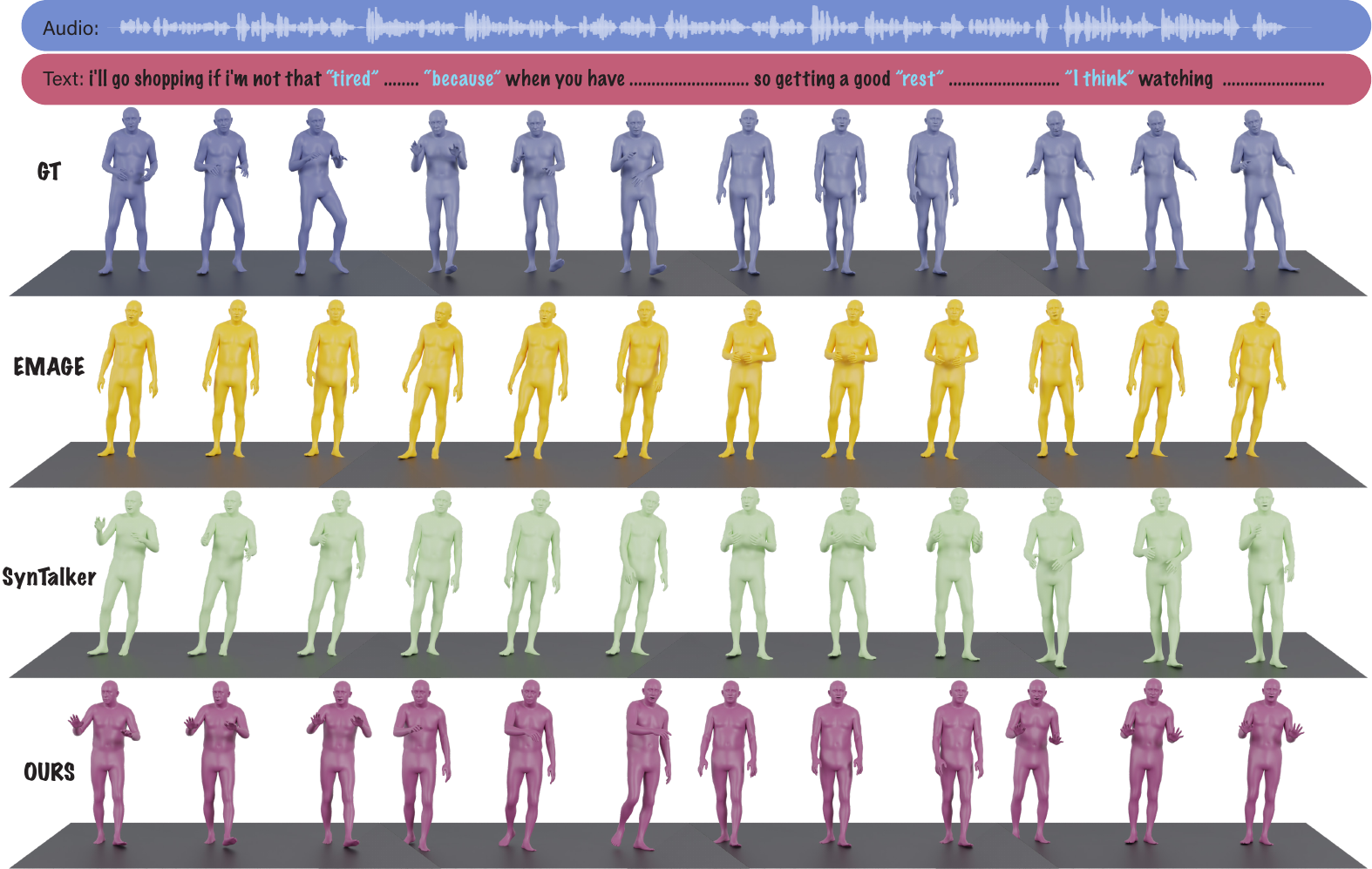}
    \vspace{-0.2in}
    \caption{Qualitative example on co-speech gesture generation. Given a speech, we visualize the ground truth 3D motion accompanying the audio, the motion generated by the baseline EMAGE~\cite{liu24emage}, SynTalker~\cite{chen2024Synerg} and our method. Our model generates more diverse and expressive motion compared to the baseline, especially when the speaker emphasizes on certain words such as ``tired" and ``because".}
    \label{fig:co-speech}
    \vspace{-0.1in}
\end{figure*}

To further understand our model's performance, we show some qualitative results of our model in Fig.~\ref{fig:co-speech}. We can see that our model generates gestures that are synchronized with the speech. See Supp. video for more examples.

\begin{figure}
    \centering
    \includegraphics[width=\linewidth]{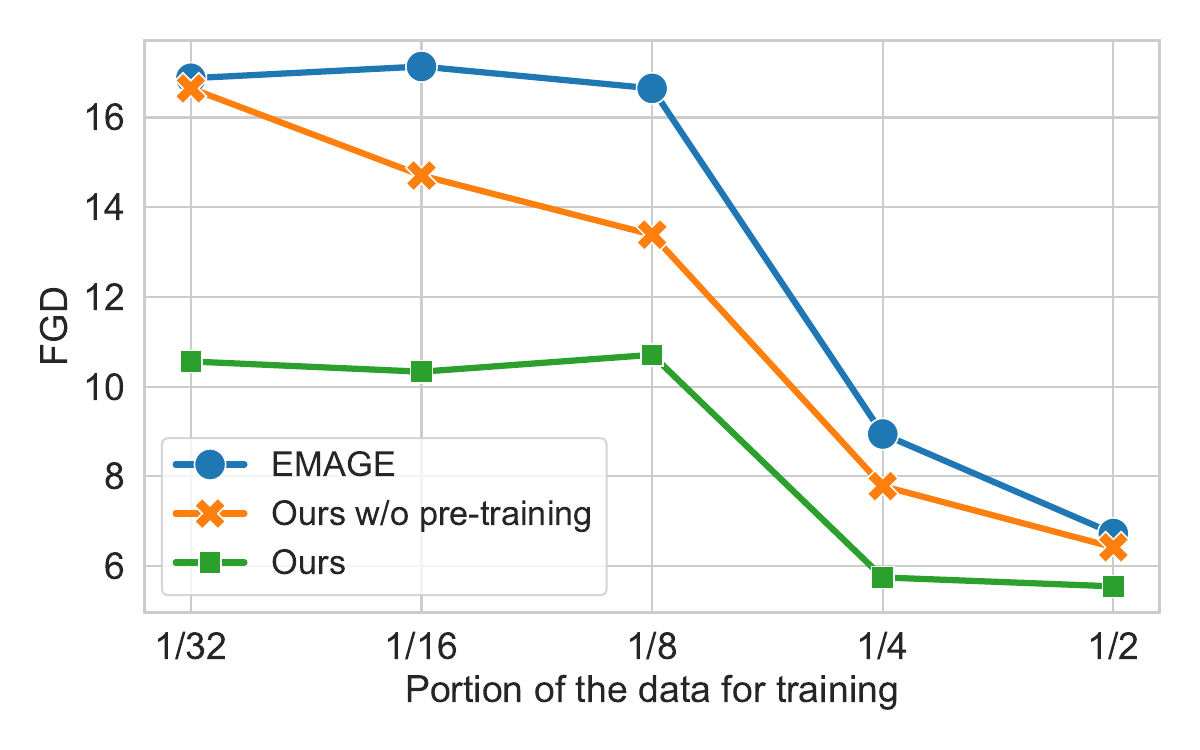}
    \vspace{-0.3in}
    \caption{Generation performance vs. the amount of post-training data. Our model learns a stronger motion prior from pre-training and thus shows much better under data scarcity.}
    \vspace{-0.1in}
    \label{fig:perf_vs_data}
\end{figure}

\subsection{Effect of Generative Pre-training}
Generating gesture motion for a new speaker requires collecting high-quality motion data, typically from motion capture systems. Collecting such data is time-consuming. In this section, we first validate the importance of each pre-training task and then investigate whether our generative pre-training leads to better generalization on new speakers and thus reduces the amount of data required for training.

\noindent\textbf{Validating pre-training tasks.} To understand how different pre-training objectives contribute to the performance, we ablate the audio-to-text alignment task (``w/o A2T"), the spatial body motion alignment task (``w/o spatial"), the temporal body motion alignment task (``w/o temporal") and the whole body alignment task (``w/o motion"). 

\begin{figure*}[th]
    \centering
    \includegraphics[width=1\textwidth]{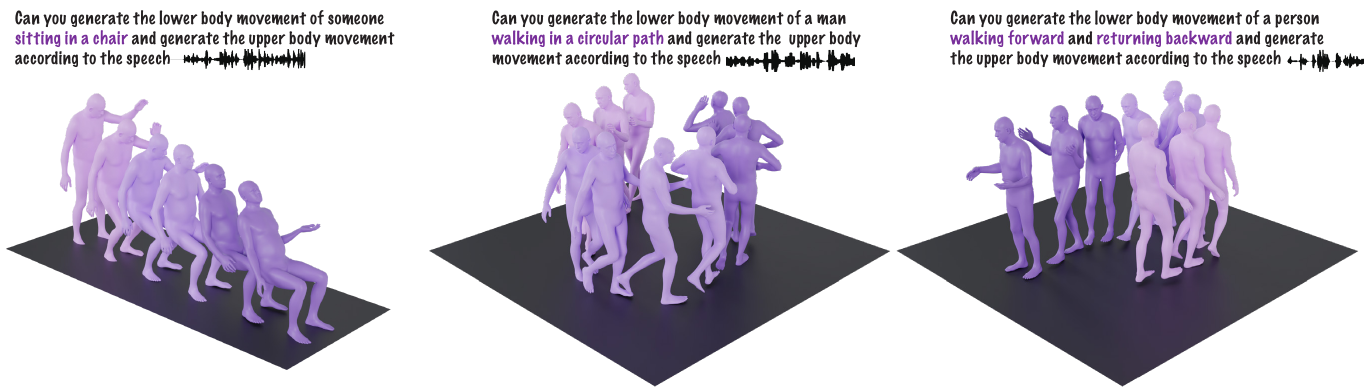}
    \vspace{-0.2in}
    \caption{Editable gesture generation. We prompt the language with text and audio information and it outputs motions that are both expressional gesture motion as well as general movement motion.}
    \label{fig:emergent}
    \vspace{-0.1in}
\end{figure*}

The results are shown in Table~\ref{tab:audio2motion}. ``w/o A2T" lowers the model's performance, indicating that aligning the audio embedding space with text helps with semantic understanding and also the downstream gesture generation task. Removing either spatial motion prediction, temporal motion prediction or them altogether hurts the performance, showing that learning spatial-temporal motion priors in the pre-training stage is important for the downstream tasks.

\noindent\textbf{Effect on the training data.} We hypothesize that our pre-training strategy captures strong multimodal correlation and motion priors, which could reduce the reliance on the amount of paired data for downstream tasks. To validate this hypothesis, we follow the setting in Sec.~\ref{sec:audio2motion} and limit the amount of training data available to the model during the pre-training stage. Note that the model has never seen audio2motion data during pre-training. We set the amount of the data to $\frac{1}{2^n}, n \in [1...5]$. We train our full model, our model without pre-training and EMAGE to convergence under each setting and evaluate on the same test set.

The results are shown in Figure~\ref{fig:perf_vs_data}. We can see that our full model starts with much lower FGD compared with the model without pre-training even when only using 1/32 of the paired training. As expected, as the amount of paired fine-tuning data increases, the performance reduces but our full model always outperforms the w/o pre-training ablation and EMAGE, showing that our model benefits pre-training and shows greater generalization under extreme data scarcity.

\subsection{Unifying Audio-to-Motion and Text-to-Motion for Editable Generation}
By taking a language model-driven approach, our model is capable of following both audio and text prompts. We first train the motion tokenizer on both BEATv2 and AMASS~\cite{mahmood2019amass} datasets since the range of motion in these two datasets is very different. We use the same tasks for pre-training. For post-training, we combine Audio2Motion and Text2Motion with various instructions, in which text-to-motion with HumanML3D~\cite{guo2022generating} text annotations. See Supp. for details.

\begin{figure}[t]
    \centering
    \vspace{-0.1in}
    \includegraphics[width=0.9\linewidth]{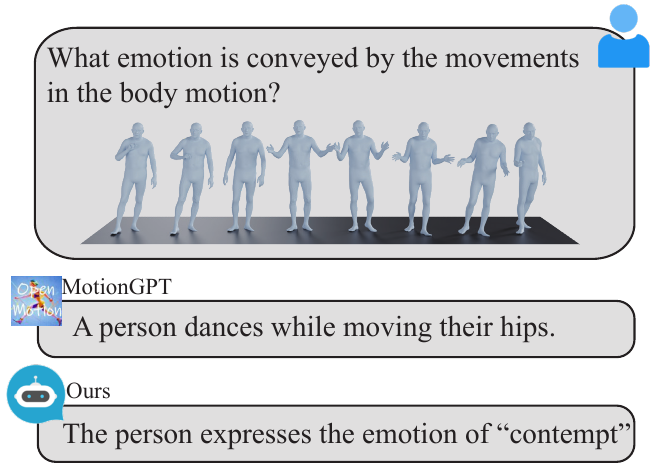} 
    \vspace{-0.1in}
    \caption{Qualitative example of emotion prediction.}
    \label{fig:motion2emotion}
    \vspace{-0.1in}
\end{figure}

By training on both text-to-motion and audio-to-motion data, our model supports joint audio-text prompts, enabling what we call editable gesture generation. This approach facilitates the generation of synergistic full-body motions conditioned on both speech and flexibly chosen prompts. For instance, the model can generate the motion of a person walking while talking. In this work, we demonstrate this capability by prompting the model separately for specific body part motions and then combining them seamlessly.
Combining conversation gestures with daily motions is extremely useful for applications such as gaming or VR. We show several qualitative examples in Fig.~\ref{fig:emergent}. We can see that the model can generate human motion that follows both audio and text prompts, showing the emergent capabilities of our model. See Supp. video for more examples.



\subsection{Predicting Emotion from Motion}
Our model's flexibility in the input/output modality also unlocks an array of new tasks such as translation between different body parts or modalities. In this section, we propose a novel task that predicts emotion from motion.

Reading someone's body language, i.e., predicting emotion from motion is important for applications such as mental health or psychiatry, however, existing audio2motion or motion2text do not have this capability. We extract the emotion labels (\textit{neutral, anger, happiness, fear, disgust, sadness, contempt, and surprise}) on BEATv2 and convert them into instructions for training. To be compatible with arbitrary language output from MotionGPT, we evaluate the model performance by measuring the BLEU~\cite{papineni2002bleu}, Rouge Cider~\cite{lin2004rouge}, and BERTScore~\cite{zhang2019bertscore} between the prediction and the ground truth, which measures the semantic distances between texts. See more details in Supp.

The results are shown in Table~\ref{tab:motion2emotion}. MotionGPT entirely fails this task with a performance similar to a random baseline because it was only trained to caption general motion rather than subtle gesture movement and body language. Our model outperforms the random and MotionGPT by a large margin, showing our model's ability to predict the emotion from motion. We also show one qualitative example in Fig.~\ref{fig:motion2emotion}.

\begin{table}
\resizebox{0.45\textwidth}{!}{
    \begin{tabular}{ccccc}
        \toprule
                & Bleu@1$\uparrow$ & Rouge Cider$\uparrow$ & BertScore$\uparrow$\\
        \midrule
        GT & 100 & 100  & 99.9\\
        Random & 2.45 & 4.44  & 0.19\\
        MotionGPT & 1.68 & 10.67  & 2.31  \\
        Ours  & \textbf{14.71} & \textbf{26.67} &  \textbf{16.94} \\
        \bottomrule
    \end{tabular} }
    \vspace{-0.1in}
    \caption{Motion to emotion. We prompt our model to predict emotion given a motion sequence.}\label{tab:motion2emotion}
    \vspace{-0.2in}
\end{table}



\section{Discussion}
In this work, we propose a novel multimodal language model to unify verbal and non-verbal language with a novel pre-training objective. Our model not only shows state-of-the-art performance on co-speech gestures but also unlocks an array of novel tasks.

While promising, the model sometimes fails to produce coherent motion potentially due to discrete motion tokenization. Moving forward, we believe incorporating continuous tokenization is an important step to improve the quality of the generated motion.

We believe unifying verbal and non-verbal language of human motion generation and understanding is crucial for real-world applications, and language models provide a powerful framework to approach that goal.

\paragraph{Acknowledgments:} This project was partially funded by NIH grant R01AG089169 and UST. The authors would also like to thank Georgios Pavlakos for his valuable discussion,  Chaitanya Patel, Jingyan Zhang, and Bin Li for their feedback on the paper.  



{
    \small
    \bibliographystyle{ieeenat_fullname}
    \bibliography{changan,specific}
}

\clearpage
\begin{table*}[t]
    \centering
    \resizebox{\linewidth}{!}
    {
    \begin{tabular}{p{4.3cm}|p{11.7cm}|p{2.1cm}}
        \toprule
         Task & Input & Output  \\
        \midrule
         Audio-to-Full Motion & \makecell[l{{p{11.7cm}}}]{Based on [audio], generate a synchronized movement sequence involving both face, hands, upper and lower body.\\ Listen to [audio] and produce movements that involve both the upper and lower body in harmony.} & {[face][hands] [upper][lower]}  \\
        \midrule
        Audio-to-Full Motion & \makecell[l{{p{11.7cm}}}]{Based on [audio], generate a synchronized movement sequence involving both face, hands, 
        upper and lower body. \\ Listen to [audio] and produce movements that involve both the upper and lower body in harmony.} & [face][hands] [upper][lower]  \\
        \midrule
        Audio\&Transcript-to-Full Motion  & \makecell[l{{p{11.7cm}}}]{Generate a set of movements for face, hand, upper, and lower body that correspond to the
        timestamped alignment in [audio\&transcript]\\
        Using the precise timestamp match in [audio\&transcript], generate corresponding face, hand, upper, and lower body movements.} &  [face][hands] [upper][lower]  \\
        \midrule
        Audio-to-Upper Body Motion  & \makecell[l{{p{11.7cm}}}]{Using [audio], produce upper body movements that capture the tone and energy. \\ From [audio], create a series of gestures that use the upper body to reflect its flow.} &  [upper]    \\
        \midrule
        Audio-to-Lower Body Motion  & \makecell[l{{p{11.7cm}}}]{Interpret [audio] with lower body gestures that reflect its tempo. \\ Create leg and foot movements that align with the intensity shifts in [audio].}  & [lower] \\
        \midrule
        Audio-to-Hands Body Motion  &  \makecell[l{{p{11.7cm}}}]{Develop a set of hand movements that respond dynamically to [audio]. \\ Generate expressive hand gestures that reflect the cues in [audio].}  & [hand]\\
        \midrule
        Audio-to-Face Body Motion  & \makecell[l{{p{11.7cm}}}]{Create expressions that correspond to the varying sentiments in [audio]. \\ Listen to [audio] and generate a sequence of facial expressions that match its energy. } & [face] \\
        \midrule
        Emotion-to-Motion & \makecell[l{{p{11.7cm}}}]{Generate a movement sequence that fully embodies the emotion of [emotion] using the face,  hands, upper body, and lower body. \\ Express the emotion [emotion] through a series of actions involving the face, hands, upper, and lower body."} & [face][hands] [upper][lower]\\
        \midrule
        Motion-to-Emotion      &  \makecell[l{{p{11.7cm}}}]{What emotion is conveyed by the movements in the face, hands, upper body, and lower body within [face][hands][upper][lower]? \\ Examine the face, hand, upper, and lower body movements in [face][hands] [upper][lower] to interpret the emotional tone.}&  [emotion] \\
        \midrule
        Text-to-Full Motion      &  \makecell[l{{p{11.7cm}}}]{Give me gestures involving the face, hands, upper body, and lower body that correspond to [caption] \\ Show me gestures involving the face, hands, upper body, and lower body that capture the essence of Input: [caption].}&  [face][hands] [upper][lower] \\
        \midrule
        Text-to-Upper Body Motion  &  \makecell[l{{p{11.7cm}}}]{Create an upper body gesture that aligns with the sentiment of [caption]. \\ Develop an upper body action sequence that mirrors the tone in [caption].}&  [upper] \\
        \midrule
        Text-to-Lower Body Motion  &  \makecell[l{{p{11.7cm}}}]{Illustrate the message in [caption] with lower body motions. \\ Translate [caption] into a lower body movement sequence.}&  [lower] \\
        \midrule
        Text-to-Lower Body Motion  &  \makecell[l{{p{11.7cm}}}]{Describe the motion represented by [upper][lower] using plain English. \\ What does the [upper][lower] communicate? Please describe it in words.}&  [caption] \\
        \bottomrule
    \end{tabular} 
    }
    \caption{Examples of instruction prompt templates during post-training. For each task, we show two examples of the input prompts and the output format.}
    \label{tab:prompt}
\end{table*}

\begin{figure*}[th]
    \centering
    \includegraphics[width=\textwidth]{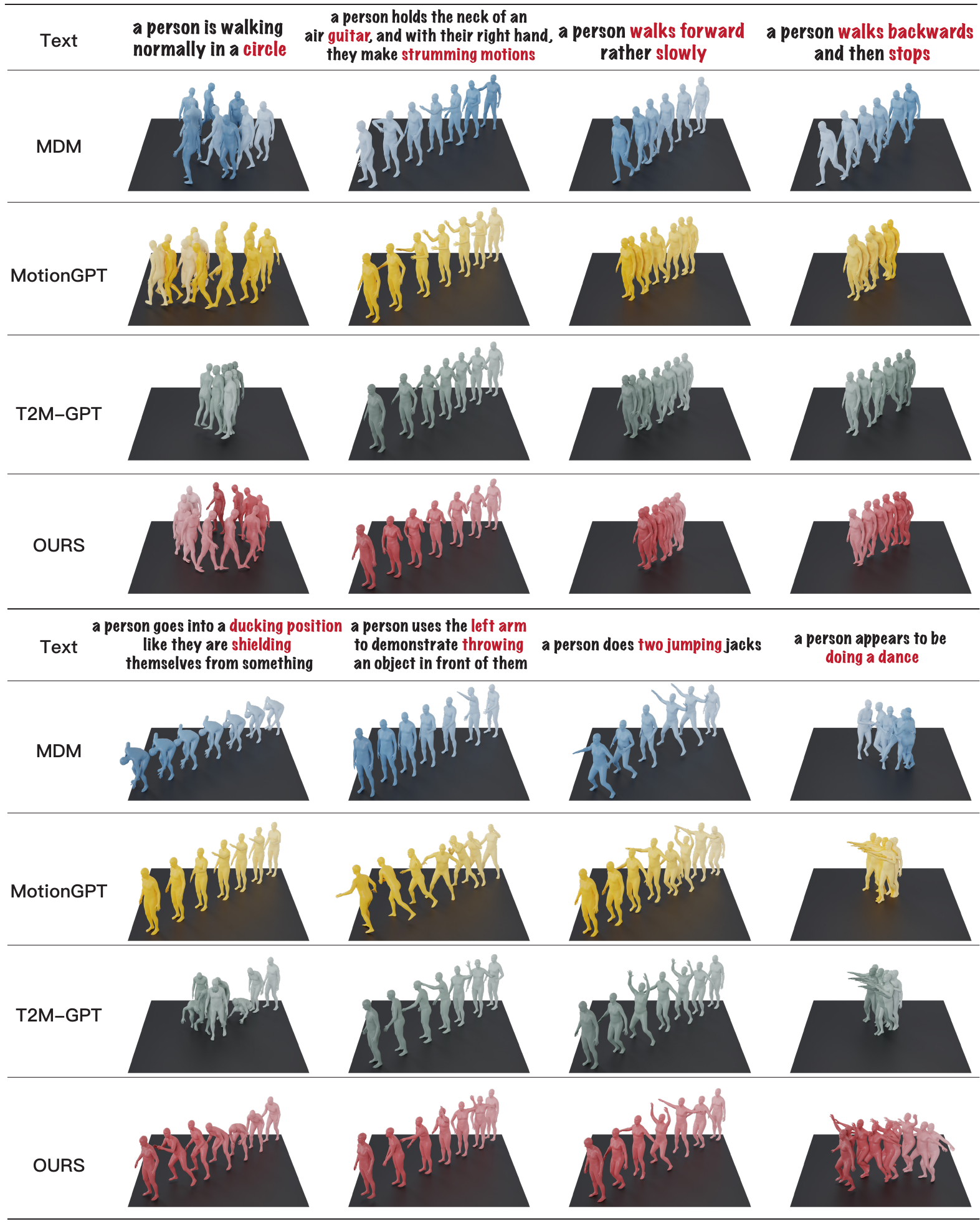}
    \vspace{-0.2in}
    \caption{Qualitative examples for text-to-motion generation. Given a text caption, we
    compare the 3D motion generated by our method with those generated by state-of-the-art methods, including MDM~\cite{tevet2023human}, T2M-GPT~\cite{zhang2023generating}, and MotionGPT~\cite{jiang23motiongpt}. Our model produces smooth, natural, and sometimes better motion in comparison with existing methods, which do not model the audio modality.}
    \label{fig:text-to-motion}
\end{figure*}

\setcounter{section}{6}
\section{Supplementary}

In this supplementary material, we provide additional details about:
\begin{enumerate}
    \item Supplementary video for qualitative examples (referenced in Sec.~1).
    \item Additional details on post-training (referenced in Sec.~3.4).
    \item Additional details on editable generation (referenced in Sec.~4.3).
    \item Additional details on emotion prediction (referenced in Sec.~4.4).
    \item Results for text-to-motion.
    \item Additional implementation details.
    \item Additional qualitative example of co-speech gesture generation.
    
\end{enumerate}

\subsection{Supplementary Video}
We provide a supplemental video to illustrate our results. In the video, we present: 1) an overview of our overall framework, 2) detailed qualitative comparisons across four tasks: co-speech gesture generation, editable gesture generation, text-to-motion generation, and emotion understanding, and 3) examples of failure cases to inspire further research. We recommend watching this video with your headphone, as video results provide a more comprehensive understanding of our approach.

\subsection{Additional Details on Post-training }

Existing datasets primarily provide pair-wise motion data but lack corresponding instructions. Following~\cite{jiang23motiongpt}, we construct paired data for downstream tasks such as co-speech gesture generation and text-to-motion generation, equipping the model with instruction-following capabilities. Building upon existing datasets~\cite{liu22beat, liu24emage, mahmood19amass, panayotov2015librispeech}, we develop an instructional multi-modal dataset comprising several core tasks. Unlike previous work~\cite{jiang23motiongpt}, our approach explicitly distinguishes each body part by introducing specific part-specific keywords. As illustrated in Tab.~\ref{tab:prompt}, each core task includes dozens of carefully designed instruction prompts.

\subsection{Additional Details on Editable Gesture Generation}

As shown in Tab.~\ref{tab:prompt}, we prompt the model with part-specific keywords, enabling it to generate any body part based on either audio or text inputs. This approach allows us to easily edit specific body parts. In this paper, we demonstrate this by prompting the model twice: once to generate the upper body from audio and once to generate the lower body from a text description. We anticipate that with further training on larger datasets, the model will be able to simultaneously follow input prompts from multiple sources.

\subsection{Additional Details on Emotion Understanding}

Since we perform instruction tuning during the post-training stage, the model does not always guarantee precise single-emotion label predictions. 
Instead of using a classification accuracy metric, we adopt text embedding distance metrics to evaluate the similarity between the predicted emotion and the ground truth labels. Specifically, we use BLEU~\cite{papineni2002bleu}, ROUGE, CIDEr~\cite{lin2004rouge}, and BERTScore~\cite{zhang2019bertscore} to assess the semantic distances between the predicted and reference texts.

\subsection{Results for Text-to-motion Generation}
In the main paper, we focused on demonstrating our model's capability in co-speech gesture generation as well as editable gesture generation. Another task that our model is naturally good at is text-to-motion generation. To understand how good our model is at generating motion from instructions, we investigate the quality of generated motion given text descriptions.

We show some qualitative examples of our text-to-motion generation in Fig.~\ref{fig:text-to-motion}, where we also compare with existing work~\cite{tevet2023human,zhang2023generating,jiang23motiongpt}. We can see that our model produces smooth, natural, and sometimes better motions in comparison with other generation methods. We encourage watching the supplementary video to get a more comprehensive understanding of our model's text-following ability.

While our model shows strong text-to-motion generation on par or even better than existing models, we observe that the common text-to-motion metrics (e.g., FID~\cite{guo2022generating}) are strongly coupled with the motion representation that existing work adopts, i.e., HumanML3D~\cite{guo2022generating} (H3D-Format), because the VAEs are trained using that format. While the H3D-Format focuses predominantly on skeletal movements, such as swinging motions, it under-represents twisting rotations and other nuanced body dynamics. In contrast, our method prioritizes expressive motion with a compositional representation, capturing a broader range of movements. Because these metrics are heavily entangled with specific motion representations, we find them not suitable to evaluate our method. We encourage readers to refer to the qualitative results in Fig.~\ref{fig:text-to-motion} and the supplementary video for a more comprehensive understanding. Future work is necessary to develop evaluation approaches that assess the quality of generated motion independently of the motion representation used.

\subsection{Additional Implementation Details}
\noindent\textbf{Model training.}
Our model employs a two-stage training process: Generative Pre-training and Post-training. During the first stage of modality alignment, we trained the full model using 8 $\times$ NVIDIA H100-80GB GPUs and the AdamW optimizer with a learning rate of 2e-4. Each configuration of the pre-trained model was trained until convergence. For the post-training stage, we used 8 $\times$ NVIDIA 3090-24G GPUs with the AdamW optimizer and a learning rate of 1e-4. To ensure fair comparisons in ablation studies, each configuration of the post-trained model was trained for a fixed 350 epochs.

\noindent\textbf{Global Translation Prediction.}
Benefiting from the compositional body representation, our approach generates high-quality expressive motions, particularly for gestures and emotion understanding. However, the holistic motion is divided into several body parts for local frames, as noted in~\cite{liu24emage}. To address this, we follow ~\cite{liu24emage} and train a VAE module with a 4-layer TCN structure. This module takes the lower body as input and estimates the global translations $ T_{trans} \in \mathbb{R}^{T\times 3} $.



\subsection{Additional Qualitative Example of Co-speech Gesture Generation}
\begin{figure*}[th]
    \centering
    \includegraphics[width=\textwidth]{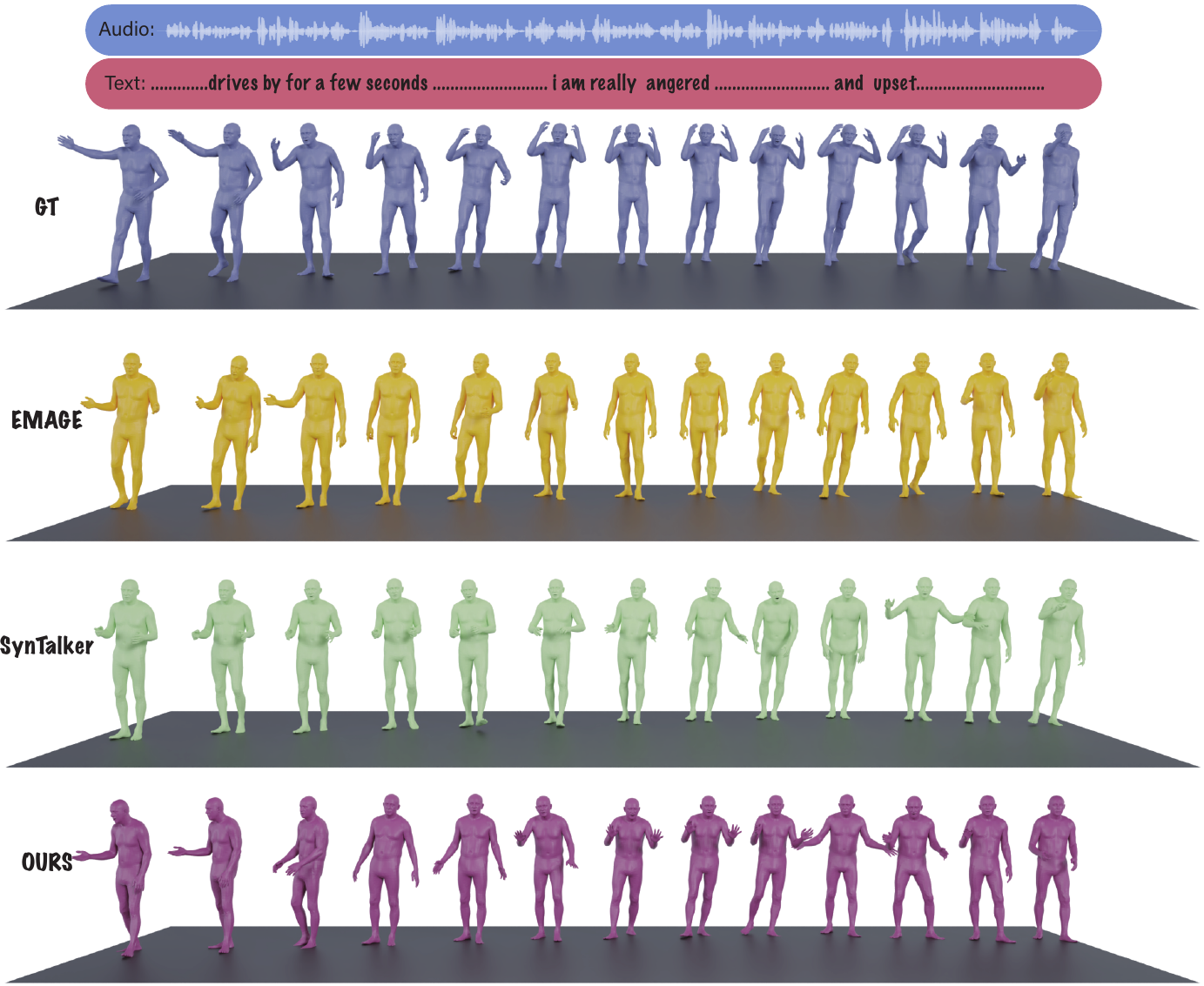}
    \caption{Additional qualitative example on co-speech gesture generation. Given an input speech, we visualize the ground truth 3D motion accompanying the audio, the motion generated by two baselines: EMAGE~\cite{liu24emage}, SynTalker~\cite{chen2024Synerg}, and our method. Our model generates more diverse and expressive motion compared to existing methods, especially when the speaker emphasizes on words such as ``angered" and ``upset".}
    \label{fig:co-speech-supp}
\end{figure*}

To show the effectiveness of our model on co-speech gesture generation, we provide one more qualitative example in Fig.~\ref{fig:co-speech-supp}. We can see that our model generates gestures that are synchronized with the speech and expressive of the emotion, outperforming two state-of-the-art methods.

\end{document}